\let\origvec\vec
\documentclass[runningheads,a4paper]{llncs}

\let\vec\origvec % Hack to avoid warning
\usepackage{amsmath}
\usepackage{amssymb}
\usepackage{xcolor}
\usepackage{graphicx}
\usepackage{bm}
\usepackage[ruled,norelsize]{algorithm2e}
\usepackage{array,booktabs,calc}
\usepackage{multirow}
\newlength{\Oldarrayrulewidth}

\usepackage{url}
\urldef{\mailsa}\path|{alfred.hofmann, ursula.barth, ingrid.haas, frank.holzwarth,|
\urldef{\mailsb}\path|anna.kramer, leonie.kunz, christine.reiss, nicole.sator,|
\urldef{\mailsc}\path|erika.siebert-cole, peter.strasser, lncs}@springer.com|
\newcommand{\keywords}[1]{\par\addvspace\baselineskip
\noindent\keywordname\enspace\ignorespaces#1}

\begin{document}

\mainmatter

\title{Accelerating likelihood optimization\\ for ICA on real signals}

\titlerunning{Accelerating likelihood optimization for ICA on real signals}

\author{Pierre Ablin\inst{1} \and Jean-Fran\c cois Cardoso\inst{2}\and Alexandre Gramfort\inst{1}}
\authorrunning{Pierre Ablin et al.}
\institute{Inria, Universit\'e Paris-Saclay, France
\and
Institut d'Astrophysique de Paris / CNRS, France}
\maketitle

\begin{abstract}
We study optimization methods for solving the maximum likelihood formulation of independent component analysis (ICA). We consider both the the problem constrained to white signals and the unconstrained problem. The Hessian of the objective function is costly to compute, which renders Newton's method impractical for large data sets. Many algorithms proposed in the literature can be rewritten as quasi-Newton methods, for which the Hessian approximation is cheap to compute. These algorithms are very fast on simulated data where the linear mixture assumption really holds. However, on real signals, we observe that their rate of convergence can be severely impaired. In this paper, we investigate the origins of this behavior, and show that the recently proposed Preconditioned ICA for Real Data (Picard) algorithm overcomes this issue on both constrained and unconstrained problems.

\end{abstract}

\section{Introduction}
Linear Independent Component Analysis (ICA)~\cite{comon1992independent} is an unsupervised data exploration technique, which models the set of observed signals as a linear instantaneous mixture of independent sources. Several methods have been proposed in the literature  for recovering the sources and mixing matrix. When formulated as a maximum likelihood estimation task, ICA becomes an optimization problem where the negative log-likelihood has to be minimized. ICA may constitute a bottleneck in practical data processing pipelines, for example due to very long signals, high number of sources or bootstrapping techniques~\cite{icasso}. It is hence crucial to maximize the likelihood as quickly as possible.

Several approaches are found in the literature. Infomax~\cite{bell1995information} can be seen as a stochastic gradient descent~\cite{cardoso1997infomax}. Several second order methods have also been proposed. In~\cite{zibulevsky2003blind}, the author propose a quasi-Newton method dubbed ``Fast Relative Newton'' method, which we will refer to as ``FR-Newton'' in the following. In~\cite{choi2007relative}, a trust-region technique is used. AMICA~\cite{palmer2012amica} also uses a quasi-Newton approach. Although it is formulated as a fixed point algorithm, FastICA~\cite{hyvarinen1999fast} is a maximum likelihood estimator under whiteness constraint of the signals~\cite{hyvarinen1999fixed}, and also behaves like a quasi-Newton method close to convergence~\cite{ablin2017ortho}.

The aforementioned algorithms all share the following property: the Hessian approximation that they use (implicitly or explicitly) stems from the expression that the true Hessian takes when the problem is solved, \textit{i.e.} when the signals are truly independent. Unfortunately, in most practical cases, the assumption that the observed signals are a mixture of independent signals is false to some extent. There might be fewer/more sources than observed signals, the sources might not be i.i.d. or stationary, they might be partially correlated, or there might be some convolutive mixture.

In the following, we demonstrate that this can lead to large differences between the true Hessian and its approximations, often leading to slow convergence on real data. We then show that the recently proposed Preconditioned ICA for Real Data (Picard) algorithm~\cite{ablin2017faster,ablin2017ortho} overcomes this problem and is able to build a better Hessian approximation.

This article is organized as follows. In section~\ref{sec:mlica}, we recall the maximum likelihood formulation of ICA, study the objective function, and derive a classical Hessian approximation. In section~\ref{sec:qnalgo}, we give some classical results about quasi-Newton algorithms, and show how the convergence speed is linked with the distance between the true Hessian and the approximation. Section~\ref{sec:picard} contains a brief description of the Picard algorithm. Finally, we illustrate the previous result with experiments in section~\ref{sec:expe}. We show that Picard builds a much better Hessian approximation that those used in previous algorithms. Through extensive experiments, we show that this leads indeed to faster convergence.

\subsubsection*{Notation}
The mean of a time-indexed sequence $x(t)_{t=1\cdot \cdot T}$ is noted $\hat{E}[x(t)] \triangleq \frac1T \sum_{t=1}^T x(t)$, and its expectation is noted $\mathbb{E}[x]$.
When $M$ is a square $N\times N$ matrix, $\exp(M)$ denotes its matrix exponential, defined as $\exp(M) \triangleq \sum_{n=0}^{\infty}\frac{M^n}{n!}$.
For two $N \times N$ matrices $M$ and $M'$, we use the Frobenius scalar product:  $\langle M \lvert M' \rangle \triangleq \sum_{i, j}M_{ij}M'_{ij}$. We denote by $\lvert \lvert M \rvert \rvert \triangleq \sqrt{\langle M \lvert M \rangle}$ the associated norm. For a fourth order tensor $H$ of size $N \times N \times N \times N$, the scalar product with respect to $H$ is defined as $\langle M \lvert H \lvert M' \rangle \triangleq \sum_{i, j, k, l}H_{ijkl}M_{ij}M'_{kl}$
The \emph{spectrum} $\text{Sp}(B)$ of a linear symmetric operator $B$
is the set of its eigenvalues.
The Kronecker symbol $\delta_{ij}$ is equal to $1$ when $i=j$ and to $0$ otherwise.

\section{Maximum-likelihood ICA}
\label{sec:mlica}
In this section, we derive the maximum-likelihood formulation of ICA, and study the underlying objective function.
\subsection{Objective function}
One observes $N$ temporal signals $x_1(t), \cdots, x_N(t)$ of $T$ samples each. The signal matrix is $X = [x_1(t), \cdots, x_N(t)]^{\top} \in \mathbb{R} ^{N \times T}$.

For the rest of this article, we assume without loss of generality that $X$ is white, \textit{i.e.} the covariance $C \triangleq \frac1T X X^{\top} = I_N$.  This can be enforced by a preprocessing whitening step: multiplying $X$ by a square root inverse of $C$.

The linear ICA model considered here is the following~\cite{comon1992independent}: there are $N$ statistically independent and identically distributed signals, $s_1(t), \cdots, s_N(t)$, which are noted as $S \in \mathbb{R} ^{N \times T}$ in matrix form, and an invertible matrix $A \in \mathbb{R} ^{N \times N}$ such that $X = AS$. The $s_i$ are referred to as sources, and $A$ is called the mixing matrix. The aim is to estimate $A$ and $S$ given $X$. In the following, $p_i$ denotes the probability density function (p.d.f.) of the $i$-th source $s_i$.

The likelihood of $A$ writes~\cite{pham1997blind}:
\begin{equation}
    p(X \lvert A) = \prod_{t = 1}^{T} \frac{1}{\lvert \det(A) \rvert} \prod_{i = 1}^{N}p_i([A^{-1} X]_{it}) \enspace .
\end{equation}
It is more practical to work with the averaged negative log-likelihood, and the variable $W = A^{-1}$ called the \emph{unmixing matrix}. In the following, $Y \triangleq WX$ denotes the current estimated sources. We define $\mathcal{L}(W) \triangleq -\frac1T \log(p(X \lvert W^{-1}))$. It writes:
\begin{equation}
    \mathcal{L}(W) = - \log\lvert \det W\rvert + \sum_{i=1}^N \hat{E}[-\log(p_i(Y_{it})) ] \enspace,
\end{equation}
where $\hat{E}$ denotes the time-averaging operation.
FastICA attempts to minimize $\mathcal{L}(W)$ under whiteness constraint $WW^{\top} = I_N$.

\subsection{Relative gradient and Hessian}
To study the variations of $\mathcal{L}$, it is convenient to work in a relative framework~\cite{cardoso1996equivariant}, where the gradient $G$ and Hessian $H$ are given by the Taylor expansion of $\mathcal{L}(\exp(\mathcal{E})W)$ where $\mathcal{E}$ is a small $N \times N$ matrix. $G$ and $H$ are implicitly defined by the equation:

\begin{equation}
\mathcal{L}(\exp(\mathcal{E})W) = \mathcal{L}(W) + \langle G \lvert \mathcal{E}\rangle + \frac{1}{2} \langle \mathcal{E} \lvert H \rvert  \mathcal{E} \rangle +  \mathcal{O}(\lvert \lvert \mathcal{E} \rvert \rvert ^3)\enspace .
\end{equation}

$G$ is a square $N \times N$ matrix, and $H$ is a linear operator from matrices to matrices, which can be seen as a $N \times N \times N \times N$ tensor. In the following, $\psi_i \triangleq -\frac{p'_i}{p_i}$ is referred to as the \emph{score function}. Simple computations yield (see~\cite{ablin2017ortho} for details):
\begin{equation}
G(W)_{ij} = \hat{E}[\psi_i(y_i)y_j] - \delta_{ij} \enspace \text{for} \enspace 1 \leq i, j \leq N
\label{eq:gradient}
\end{equation}
\begin{equation}
H(W)_{ijkl} = \delta_{i l} \delta_{j k} \hat{E}[\psi_i(y_i)y_i]  + \delta_{ik} \, \hat{E}[\psi_i'(y_i)y_jy_l]  \enspace \text{for} \enspace 1 \leq i, j, k, l \leq N
\end{equation}
The Hessian is sparse since it has of the order of $N^3$ non-zero coefficients. Still, its evaluation requires computing  $O(N^3)$ sample averages $\hat{E}[\psi_i'(y_i)y_jy_l]$, making the standard Newton's method impractical for large data sets.

\subsection{The Hessian approximation}

If the signals $(y_1(t),\cdots, y_N(t))$ are independent, then $\mathbb{E}[\psi_i'(y_i)y_jy_l] = \delta_{jl}\mathbb{E}[\psi_i'(y_i)y_j^2]$. A natural approximation of $H$ is then :
\begin{equation}
œ = \delta_{i l} \delta_{j k} \hat{E}[\psi_i(y_i)y_i]  + \delta_{ik} \delta_{jl}\, \hat{E}[\psi_i'(y_i)y_j^2] \enspace .
\end{equation}
This approximation matches the true Hessian \textbf{if the number of samples goes to infinity and the $(y_i)$ are independent}. If the linear ICA model holds, i.e. if there exists independent signals $S$ and a mixing matrix $A$ such that $X  =AS$, then, for $W^* = A^{-1}$, $\tilde{H}(W^*) = H(W ^*) + \mathcal{O}(\frac{1}{\sqrt{T}})$. As the number of samples is generally large, the approximation is very good in that case.

However, in a practical case, ICA is performed on real data for which the ICA model does not hold exactly.  In that case, even for $W^* = \arg\min \mathcal{L}(W)$, one does not necessarily have $\mathbb{E}[\psi_i'(y_i)y_jy_l] = \delta_{jl}\mathbb{E}[\psi_i'(y_i)y_j^2]$, and $\tilde{H}(W ^*)$ may be quite far from $H(W ^*)$.

\section{Speed of convergence of quasi-Newton methods}
\label{sec:qnalgo}
In the following, we consider a general relative quasi-Newton method to minimize $\mathcal{L}$, described in algorithm~\ref{algo:quasi-newton}. It takes as input the set of mixed signals $X$, which are assumed white for simplicity, and a boolean "whiteness constraint" which determines if the algorithm works under whiteness constraint.  Note that the policy to compute the approximation $\hat{H}$ is not specified: one could use $\hat{H} = \tilde{H}$, but other choices are possible. To keep the analysis simple, we assume that the line-search is perfect, i.e. that the objective function is always minimized in the search direction.

\begin{algorithm}[tb]
\SetKwInOut{Input}{input}
\SetKwInOut{Output}{output}
\SetKwBlock{Repeat}{repeat}{}
 \Input{Set of white mixed signals $X$, boolean ``whiteness constraint''}
 Set $W = I_N$ \;
 Set $Y = X$ \;
 \Repeat{
 Compute the gradient $G$ using~\eqref{eq:gradient}\;
  \If{whiteness constraint}{
 Project $G$ on the antisymmetric matrices: $G\leftarrow \frac12(G - G^{\top})$\;
 }
 Compute a Hessian approximation $\hat{H}$ \;
 Compute the search direction $D = - \hat{H}^{-1} G$ \;
 \If{whiteness constraint}{
 Project $D$ on the antisymmetric matrices: $D\leftarrow \frac12(D - D^{\top})$\;
 }
 Compute the step size $\alpha = \arg\min_{\alpha} \mathcal{L}(\exp(\alpha D) W)$ using line-search \;
 Set $W \leftarrow \exp(\alpha D) W$ \;
 Set $Y = WX$ \;
 }
 \Output{Unmixing matrix $W$, unmixed signals $Y$.}
 \caption{Quasi-Newton method for likelihood optimization}
 \label{algo:quasi-newton}
\end{algorithm}

\subsection{Theoretical results}
\label{sec:theory}
Let us recall some results on the convergence speed of such method. These results mostly come from Numerical Optimization~\cite{nocedal1999optim}, chapter 3.3.

First, the following theorem shows that under mild assumptions, the sequence of unmixing matrices produced by algorithm~\ref{algo:quasi-newton} converges to a local minimum of~$\mathcal{L}$.

\begin{theorem}
Assume that the sequence of Hessian approximations $\hat{H}$ used in algorithm~\ref{algo:quasi-newton} is positive definite, of spectrum lower bounded by some constant $\lambda_{min}>0$. Then, the sequence of unmixing matrices generated by the algorithm converges towards a matrix $W^*$ such that $G(W ^*)=0$ and $H(W^*)$ is positive definite.
\label{theo:conv}
\end{theorem}

This theorem is a direct consequence of Zoutendijk's result (see~ \cite{nocedal1999optim}, theorem 3.2). Interestingly, it implies that the algorithm cannot converge to a saddle point (where $H(W^*)$ is not positive), but only towards local minima, as guaranteed for gradient based methods.

Quasi-Newton methods typically aim at finding a direction close to Newton's direction $- H^{-1}G$, and ideally have the same quadratic convergence rate.
By Theorem 3.6 in~\cite{nocedal1999optim}, this happens if and only if at convergence, the Hessian approximation matches the true Hessian in the search direction. As we have seen before, even when the ICA model holds, the simple approximation $\tilde{H}$ only matches asymptotically the true Hessian, meaning that the above theorem never practically applies. Thus, the convergence of algorithm~\ref{algo:quasi-newton} can only be linear. The following algorithm gives the rate of convergence.

\begin{theorem}
Assume that the condition of theorem~\ref{theo:conv} holds. Assume that the sequence of approximate Hessians $\hat{H}$ converges towards $\hat{H^*}$. Let $\lambda_m$ (resp. $\lambda_M$) be the smallest (resp. largest) eigenvalue of $\hat{H^*}^{-\frac12} H \hat{H^*}^{-\frac12}$ and define the condition number:
\begin{equation}
\kappa \triangleq \frac{\lambda_M }{\lambda_m} \enspace .
\end{equation}
Then, for all $ r < \frac{1}{\kappa}$ and $n$ large enough,
the sequence $W_n$ of unmixing matrices produced by algorithm~\ref{algo:quasi-newton} satisfies
$\mathcal{ L}(W_{n+1}) - \mathcal{ L}(W^*)\leq (1 - r) [\mathcal{ L}(W_{n}) - \mathcal{ L}(W^*)]$.
\label{theo:convspeed}
\end{theorem}
We now give a brief sketch of proof.

\begin{proof}
For simplicity, the proof is made in a non-relative framework, where the update rule is $W_{n+1} = W_n - \alpha \hat{H}_n^{-1}\nabla \mathcal{L}(W_n)$. First, we make the useful change of variable $U_n = \hat{H^*}̂^{\frac12}W_n$, and define the new objective function $L(U_n) = \mathcal{L}(\hat{H^*}^{-\frac12}U_n)$. Simple computations show that $U_n$ verifies $U_{n+1} = U_n - \alpha B_n \nabla L(U_n)$, where $B_n \triangleq \hat{H^*}̂^{\frac12} \hat{H}_n^{-1} \hat{H^*}̂^{\frac12}$. This sequence tends towards identity, meaning that the behavior of $U_n$ is asymptotically the same as a gradient descent. One has $\nabla ^2 L(U) = \hat{H^*}̂^{-\frac12} [\nabla ^2 \mathcal{L}(W)]\hat{H^*}̂^{-\frac12}$.

Let $\varepsilon>0$ be a small number. Since $\text{Sp}(B_n) \rightarrow \{1\}$ and $\text{Sp}(\nabla^2L(U_n) \subset [\lambda_m, \lambda_M]$ as $n$ goes to infinity, for $n$ large enough we have that $\text{Sp}(B_n) \subset [1 - \varepsilon, 1 + \varepsilon]$ and $\text{Sp}(\nabla ^2 L(U_n)) \subset [(1 - \varepsilon)\lambda_m, (1 + \varepsilon)\lambda_M]$. This means that the iterates $U_n$ are in a set where $L$ is $(1 + \varepsilon)\lambda_M-$smooth and $(1 - \varepsilon)\lambda_m-$strongly convex.
The smoothness implies the following convexity inequality:

\begin{equation}
L(V) \leq L(U) + \langle \nabla L(U)\lvert V-U \rangle + \frac{(1 + \varepsilon)\lambda_M}{2} \lvert \lvert U - V \rvert \rvert ^2
\label{eq:smoothness}
\end{equation}
and the strong convexity enforces the Polyak-Lojasiewicz conditions~\cite{karimi2016linear}:
\begin{equation}
\frac12 \lvert \lvert \nabla f(U) \rvert \rvert ^2 \geq (1 -  \varepsilon)\lambda_m[L(U) - L(U^*)]
\label{eq:polyak}
\end{equation}

Let $\beta$ be a positive scalar. For an exact line-search, we have $L(U_{n+1}) \leq L(U_n - \beta B_n \nabla L(U_n))$. Using $U = U_n$ and $V = U_n - \beta B_n \nabla L(U_n)$ in inequality~\eqref{eq:smoothness}, we obtain:
\begin{equation}
L(U_{n+1}) -L(U_n) \leq -\beta \langle \nabla L(U_n) \lvert B_n \nabla L(U_n) \rangle + \beta^2 \frac{(1 + \varepsilon)\lambda_M}{2}\lvert \lvert B_n \nabla L(U_n) \rvert \rvert ^2
\label{eq:firstineq}
\end{equation}
The condition on the spectrum of $B_n$ implies $\langle \nabla L(U_n) \lvert B_n \nabla L(U_n) \rangle \geq (1-\varepsilon) \lvert \lvert \nabla L(U_n) \rvert \rvert ^2$ and $\lvert \lvert B_n \nabla L(U_n) \rvert \rvert ^2 \leq (1+\varepsilon)^2\lvert \lvert  \nabla L(U_n) \rvert \rvert ^2$. Replacing in eq.~\eqref{eq:firstineq} yields:
\begin{equation}
L(U_{n+1}) -L(U_n) \leq \left(-\beta (1 - \varepsilon) + \beta^2 \frac{(1 +\varepsilon)^3\lambda_M}{2}\right)\lvert \lvert  \nabla L(U_n) \rvert \rvert ^2
\end{equation}
This holds for any $\beta$, in particular for $\beta = \frac{1-\varepsilon}{(1+\varepsilon)^3 \lambda_M}$ (which minimizes the scalar factor in front of $\lvert \lvert  \nabla L(U_n) \rvert \rvert ^2$). We obtain:
\begin{equation}
L(U_{n+1}) -L(U_n) \leq -\frac{(1 - \varepsilon)^2}{2(1+\varepsilon)^3 \lambda_M} \lvert \lvert  \nabla L(U_n) \rvert \rvert ^2
\end{equation}
Using eq.~\eqref{eq:polyak} then gives:
\begin{equation}
L(U_{n+1}) -L(U_n) \leq -\frac{(1 - \varepsilon)^3 \lambda_m}{(1+\varepsilon)^3 \lambda_M} [L(U_n) - L(U^*)]
\end{equation}
Rearranging the terms, we obtain the desired result for $r = (\frac{1 - \varepsilon}{1+\varepsilon})^3\frac{1}{\kappa}$.
\end{proof}

\subsection{Link with maximum likelihood ICA}

There are many ICA algorithms closely related to the minimization of $\mathcal{L}$ and similar to Algorithm~\ref{algo:quasi-newton} . For instance, Infomax is a stochastic version of algorithm~\ref{algo:quasi-newton} without whiteness constraint and with $\hat{H} = Id$. In~\cite{zibulevsky2003blind}, the author proposes to use $\hat{H} = \tilde{H}$ in algorithm~\ref{algo:quasi-newton}, without the whiteness constraint. The algorithm is denoted as ``Fast Relative Newton method'', or FR-Newton for short. The same approach is used in AMICA~\cite{palmer2012amica}.  In~\cite{ablin2017ortho}, it is shown that close to convergence, FastICA's iterations are similar to those of algorithm~\ref{algo:quasi-newton} with the whiteness constraint, and where the Hessian approximation has the same properties as  $\tilde{H}$: it coincides asymptotically with $H$ when the underlying signals $(y_i)$ are independent, but may differ otherwise. Thus, the previous results apply for a wide range of popular ICA methods.

\section{Preconditioned ICA for Real Data}
\label{sec:picard}

Let us now introduce the Preconditioned ICA for Real Data (Picard) algorithm, which finds a better Hessian approximation than $\tilde{H}$.
The algorithm is an adaptation of the L-BFGS algorithm~\cite{nocedal1980updating}. It has a memory of size $m$ which stores the $m$ previous iterates $W$ and gradients $G$. From these values, it recursively builds a Hessian approximation starting from $\tilde{H}$. In the following, $H_P$ denotes that approximation. It does so in an uninformed fashion, without any prior on the local geometry. L-BFGS has been shown to perform well on a wide variety of problems. Here, we have the advantage of having $\tilde{H}$ as a good initialization for the approximate Hessian. Another asset of this method is that the Hessian approximation never has to be computed, because there is an efficient way of computing the direction $- H_P^{-1}G$. Picard can handle both constrained and unconstrained problems. For further details for the practical implementation, see~\cite{ablin2017faster,ablin2017ortho}.

Python and Matlab/Octave code for Picard is available online.\footnote{https://github.com/pierreablin/picard}

\section{Experiments}
\label{sec:expe}

\subsection{Comparison of the condition numbers}
\begin{figure}[htp]
  \centering
  \includegraphics[width=0.49\columnwidth]{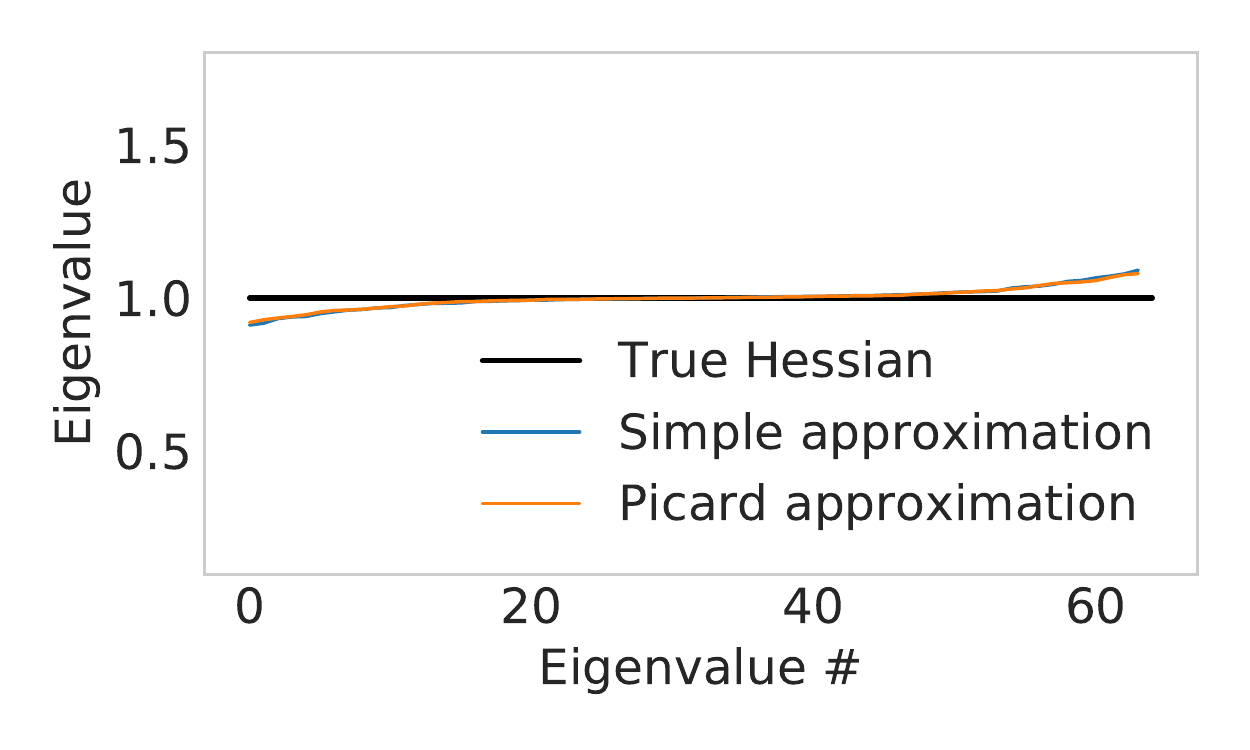}\hfill
  \includegraphics[width=0.49\columnwidth]{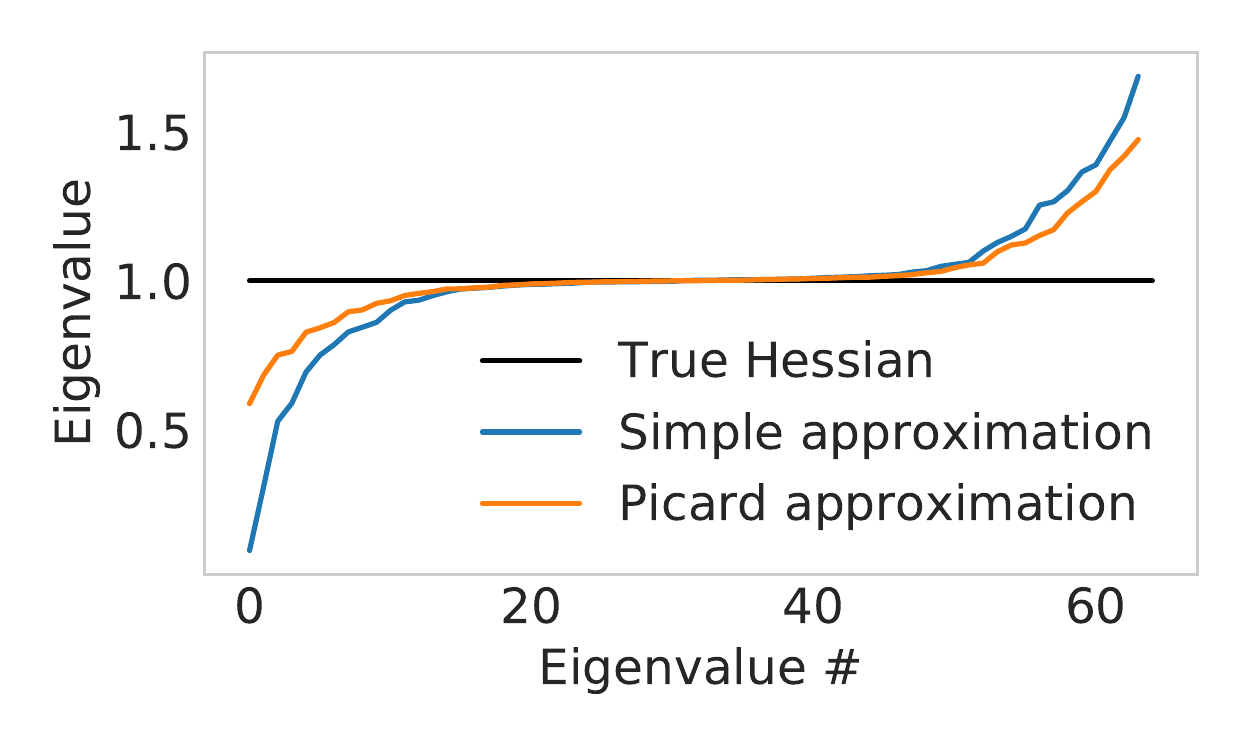}
  \caption{A measure of the closeness of the approximate Hessians to the true Hessian at the maximum likelihood: sorted spectrum of $\hat{H}^{-\frac12}H \hat{H}^{-\frac12}$. Left: simulated data where the ICA model holds. Right: real data. On the simulated data, we find $\kappa = 1.2$ for both $\hat{H} = \tilde{H}$ and $\hat{H} = H_P$. For that example on real data, we find $\kappa = 29 $ for $\tilde{H}$  and a significantly smaller $\kappa = 2.6$ for $H_P$.}
  \label{fig:diffhessian}
\end{figure}
In this section, we show how close the Hessian approximations $\tilde{H}$ and $H_P$ are to $H$ on simulated and real data.
We consider two different datasets $X$ of $N=8$ signals of length $T = 20000$. The first one is obtained by simulating a source matrix $S$ of independent signals, and a random mixing matrix $A$. We take $X = AS$. For that dataset, the linear ICA model holds by construction. The second one is obtained by extracting $20000$ square patches of size $(8,8)$ from a natural image. PCA is then applied to reduce to $8$ the number of signals.

First, we find a local minimum  $W^*$ of $\mathcal{L}(W)$
by running one of the algorithms on this dataset.
Then, the simple approximation $\tilde{H}(W^*)$, the Picard approximation $H_P(W ^*)$ and the true Hessian $H(W^*)$ are computed. As explained by theorem~\ref{theo:convspeed}, what drives the convergence speed of the algorithms is the spectrum of $\hat{H}^{-\frac12}H \hat{H}^{-\frac12}$ where $\hat{H}$ is the approximation. Figure~\ref{fig:diffhessian} displays these spectrum for the two datasets.

We observe that $H_P$ and $\tilde{H}$ are very similar on the simulated dataset, and that the resulting condition numbers are close to 1, which explains the fast convergence of the two algorithms. On the real dataset, the results are different: the spectrum obtained with $H_P$ is flatter than the one obtained with $\tilde{H}$, which means that Picard builds a Hessian approximation which is  significantly better than $\tilde{H}$.

\subsection{Convergence speed on real datasets}

We now compare the convergence speed of Picard / Picard-O with FR-Newton from~\cite{zibulevsky2003blind} and FastICA~\cite{hyvarinen1999fixed} on three types of data on which ICA is widely used.

\begin{figure}
  \begin{tabular}{ b{0.04\textwidth} m{0.04\textwidth} | m{0.32\textwidth}  m{0.32\textwidth} m{0.32\textwidth} }

  \multirow{2}{*}[-5ex]{\rotatebox[origin=c]{90}{Unconstrained}}
    & \rotatebox{90}{Iterations} &
    \includegraphics[width=0.33\columnwidth]{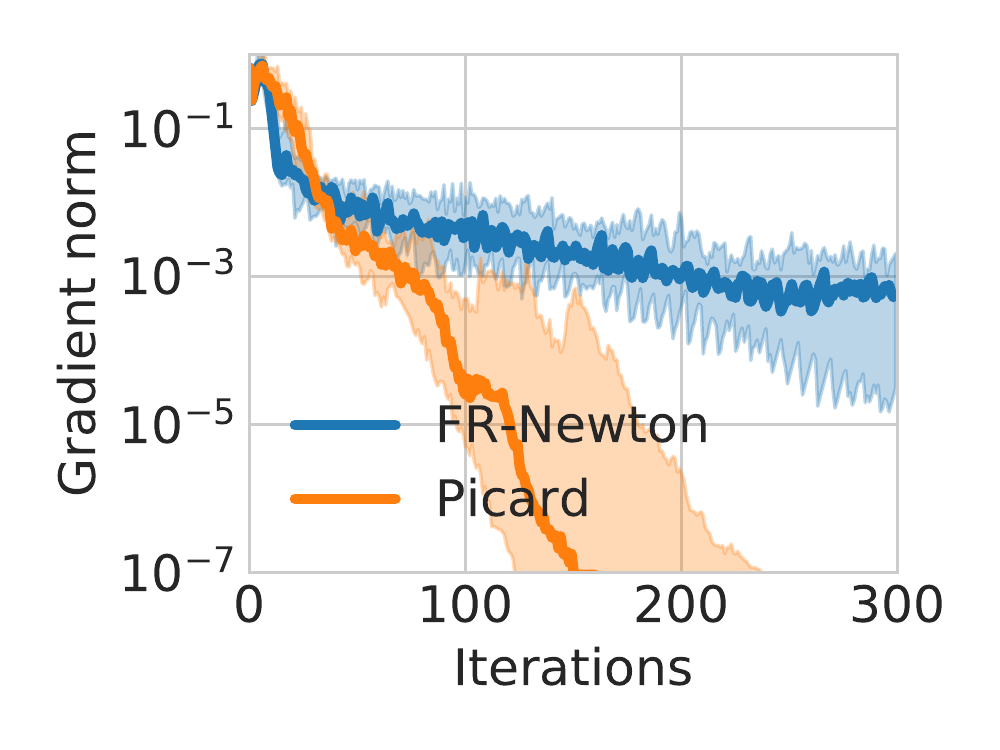}&
    \includegraphics[width=0.33\columnwidth]{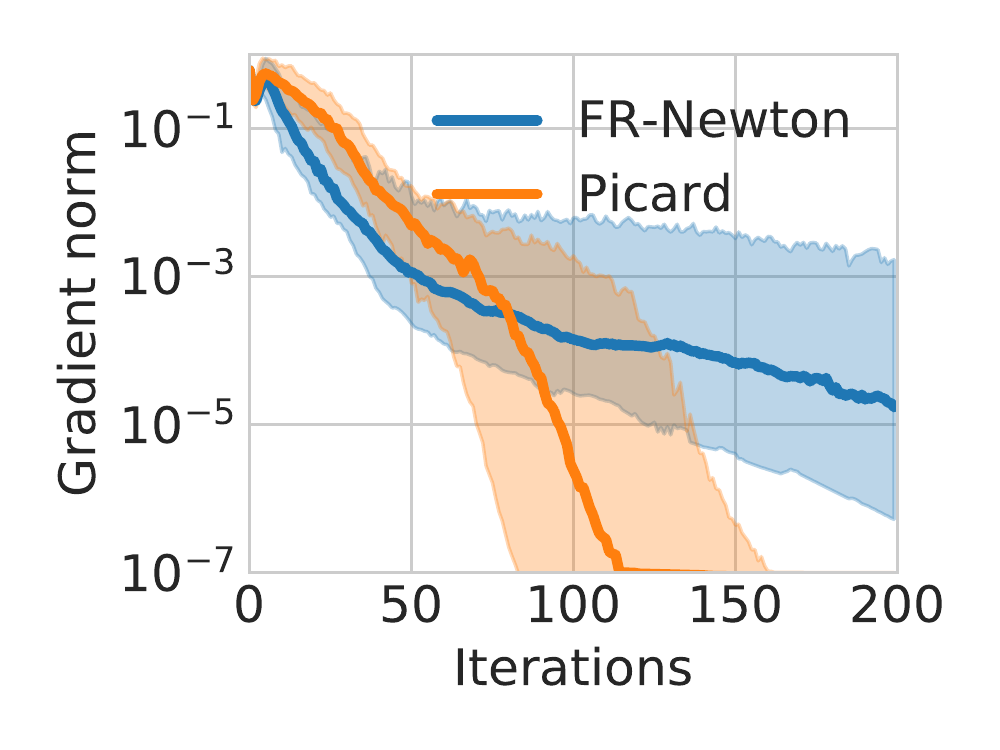}&
    \includegraphics[width=0.33\columnwidth]{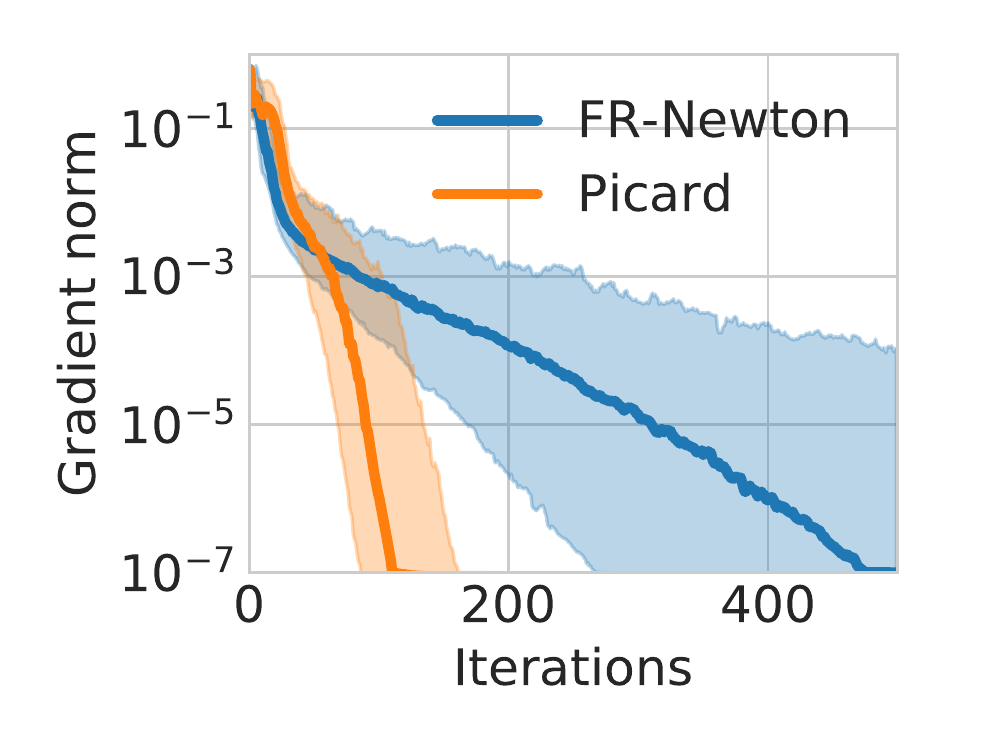}\\ \cmidrule[1pt]{2-5}
    & \rotatebox{90}{Time} &
    \includegraphics[width=0.33\columnwidth]{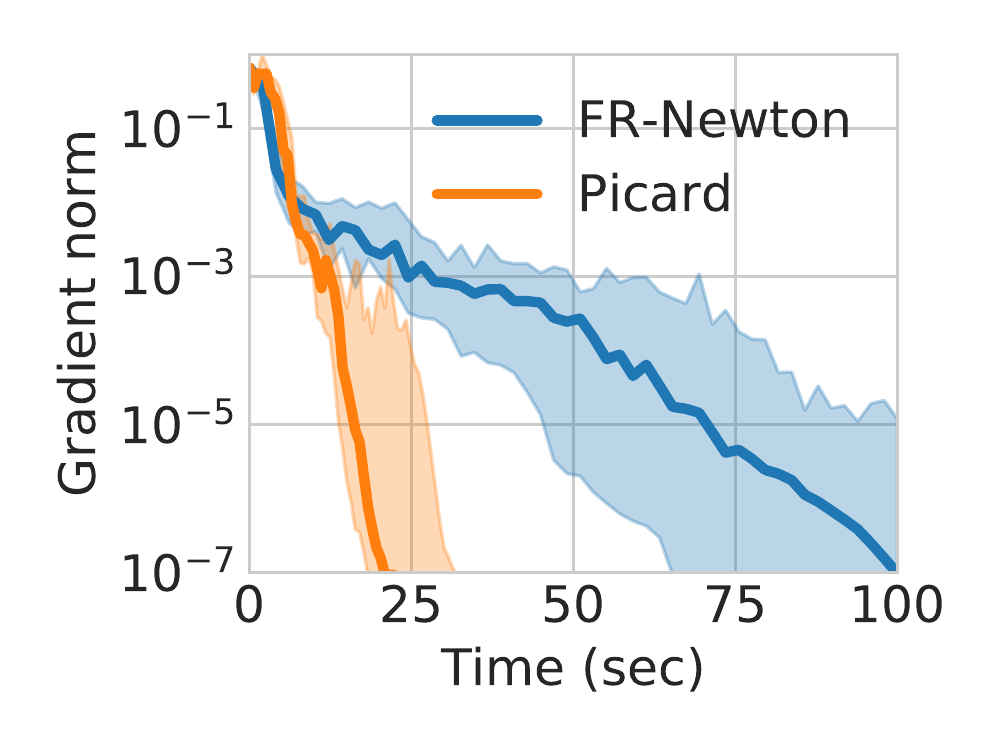}&
    \includegraphics[width=0.33\columnwidth]{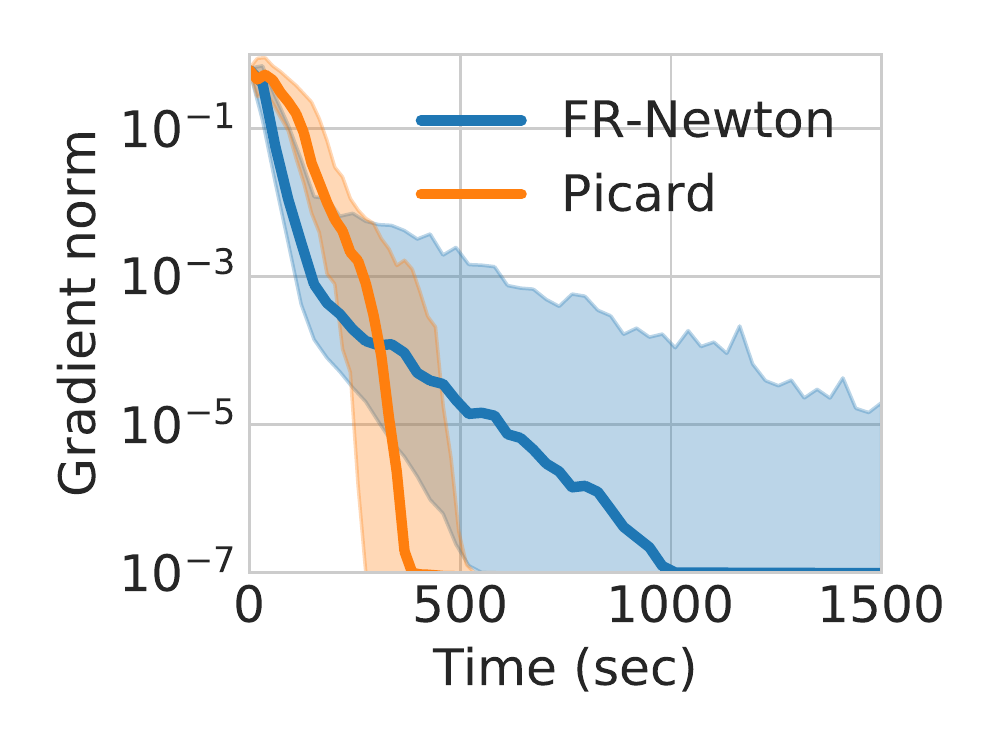}&
    \includegraphics[width=0.33\columnwidth]{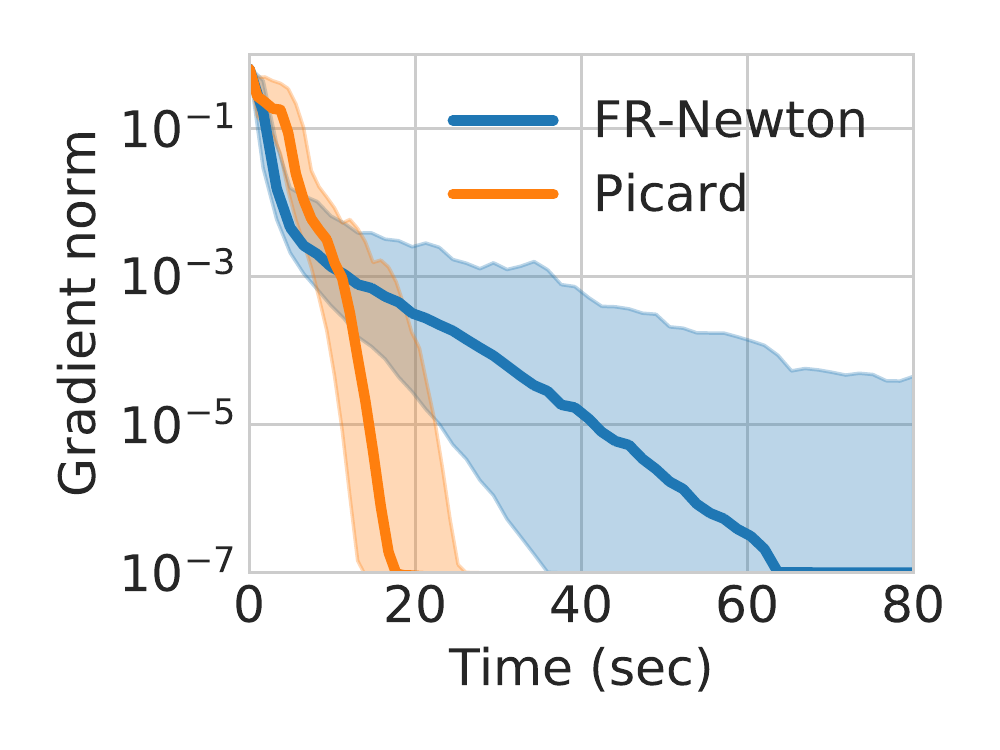}\\ \hline
    \multirow{2}{*}[-7ex]{\rotatebox[origin=c]{90}{Constrained}}
     & \rotatebox{90}{Iterations} &
    \includegraphics[width=0.33\columnwidth]{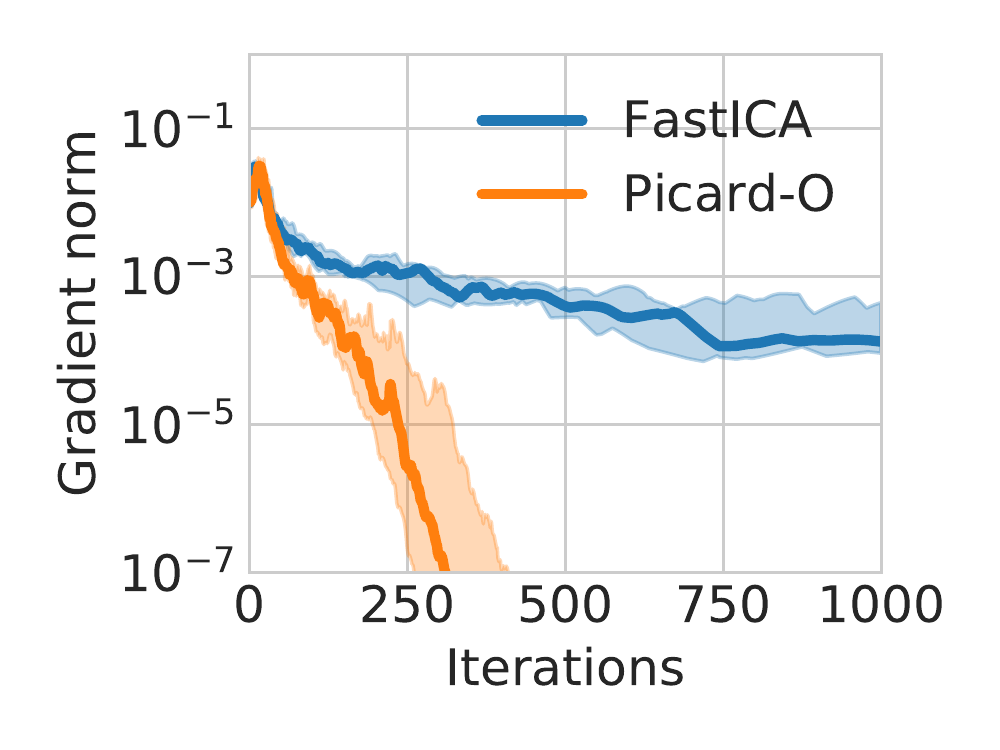}&
    \includegraphics[width=0.33\columnwidth]{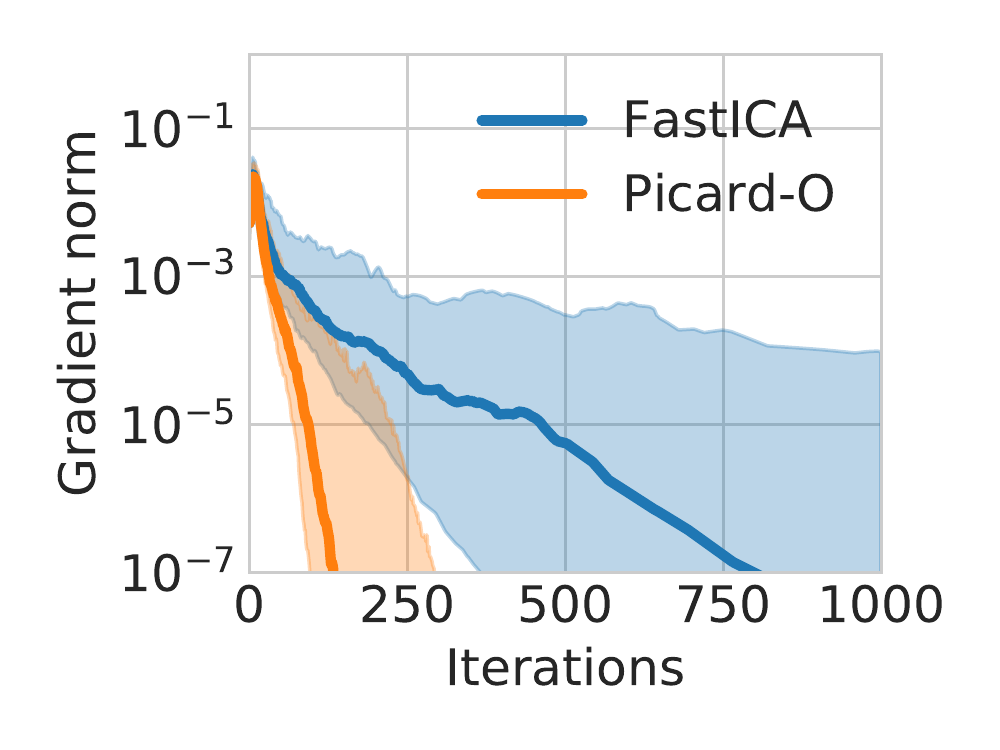}&
    \includegraphics[width=0.33\columnwidth]{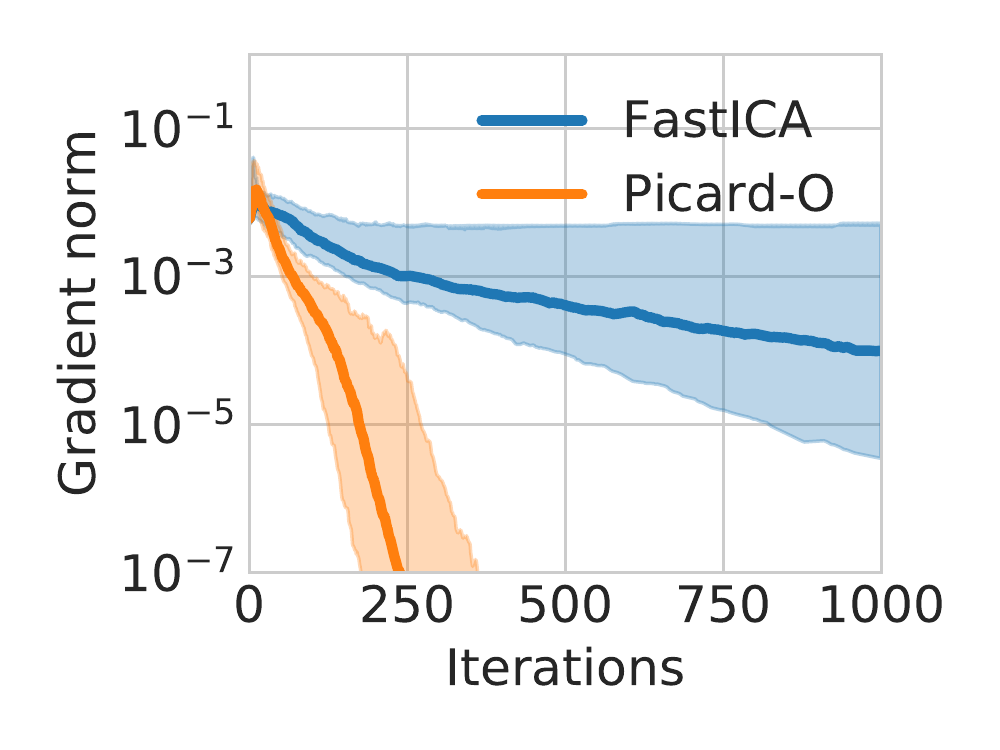}\\
     \cmidrule[1pt]{2-5}
    & \rotatebox{90}{Time} &
    \includegraphics[width=0.33\columnwidth]{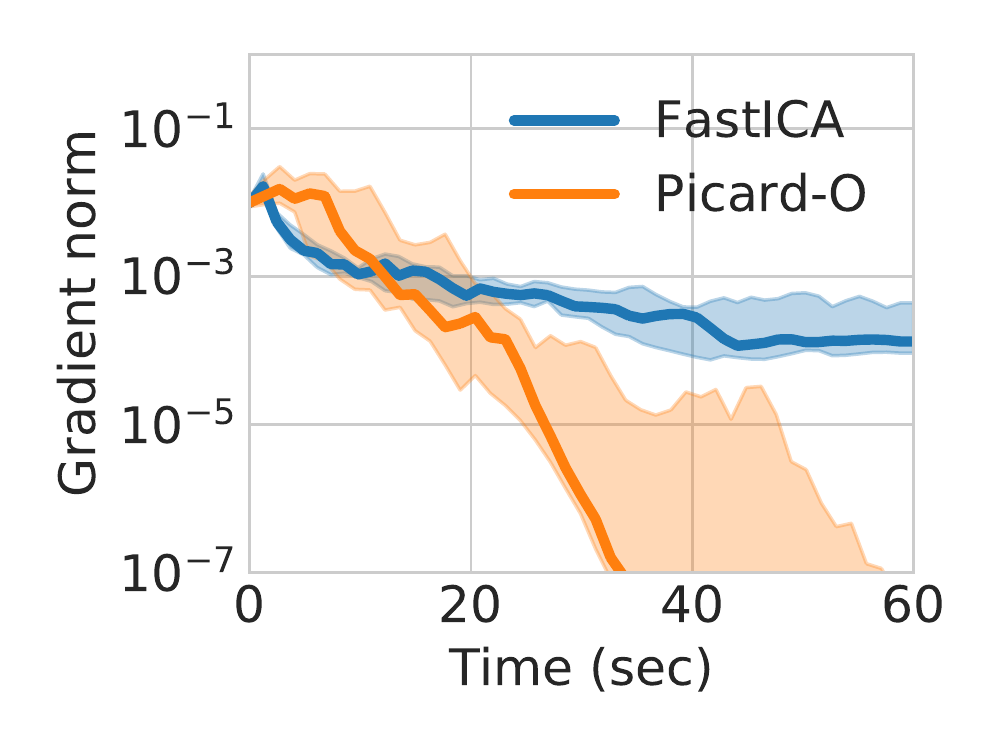}&
    \includegraphics[width=0.33\columnwidth]{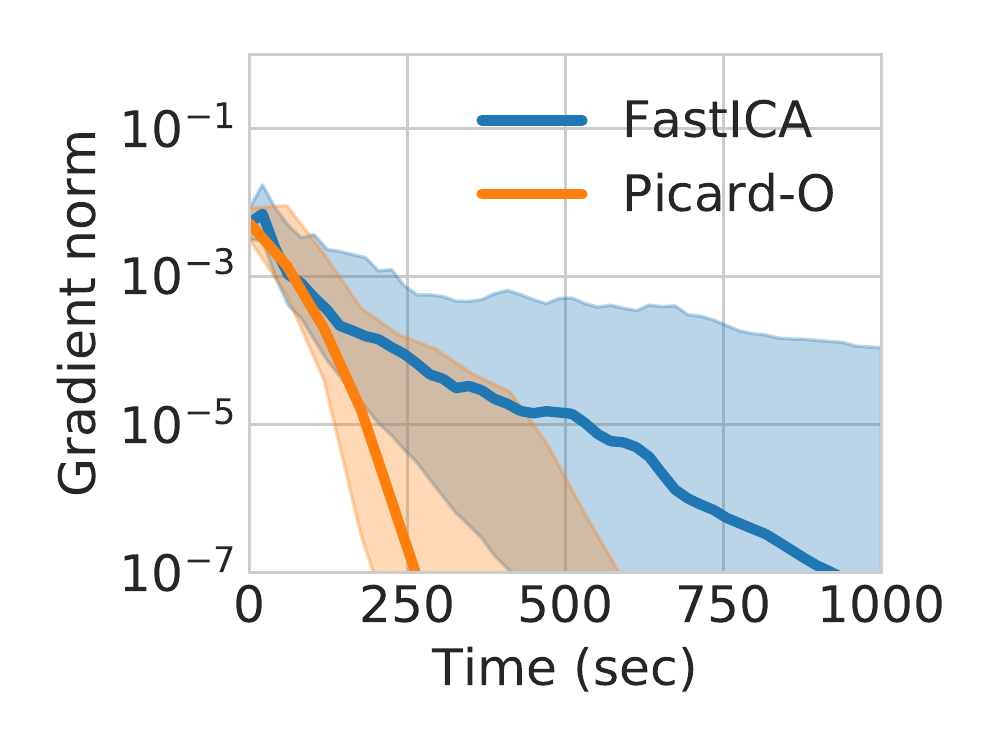}&
    \includegraphics[width=0.33\columnwidth]{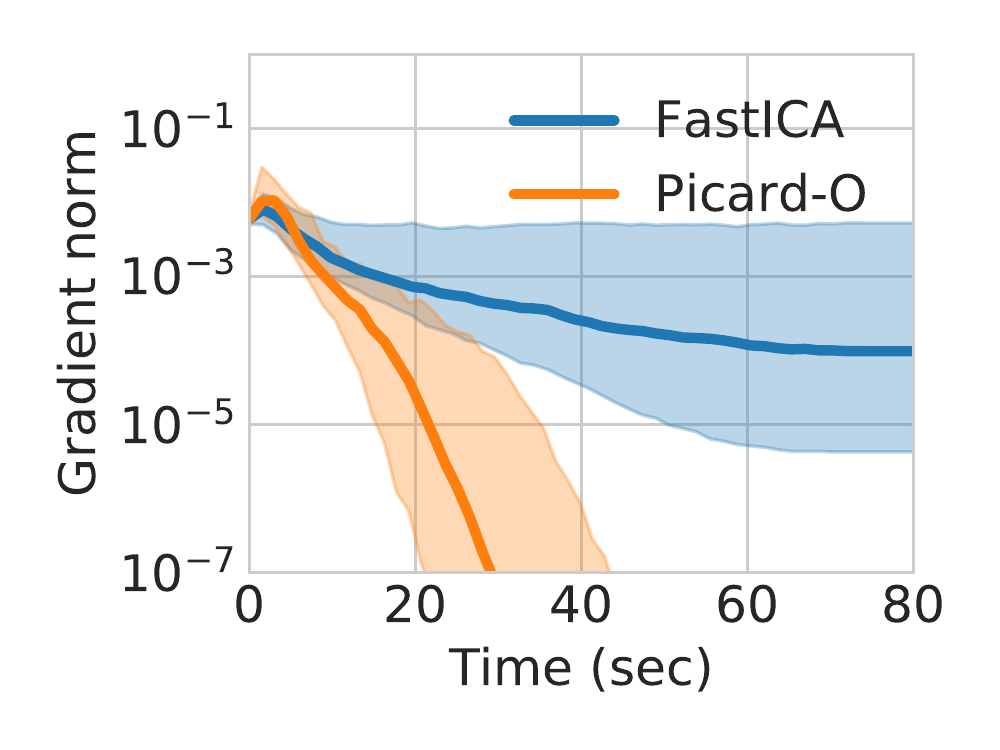}\\
    & & \centering Genomics, 10 runs & \centering EEG, 130 runs & \centering Images, 300 runs
  \end{tabular}
\caption{Convergence speed of several ICA algorithms on 3 real data sets.
Each column corresponds to a type of data. The first two rows correspond to the unconstrained algorithms, the last two to the constrained algorithms. The first row of each pair displays the evolution of gradient across iterations, the second one displays the evolution of gradient against time. Bold lines correspond to the medians of the gradient norms, and the shading displays the $10-90 \%$ percentile.}
\label{fig:res}
\end{figure}

The first is a cancer genomics dataset generated by the TCGA Research Network: \url{http://cancergenome.nih.gov}, of initial size $N \simeq 2000$ and $T \simeq 20000$ for which the dimension has been reduced to $N = 60$ by PCA. The second consists of 13 EEG recordings datasets~\cite{delorme2012independent} of size $N = 71$ and $T \simeq 300000$. The last one is 30 datasets of $T = 20000$ extracted image patches of size $(8, 8)$, flattened to obtain $N = 64$ signals. We run the aforementioned algorithms $10$ times on each datasets. We keep track of the evolution of the gradient norm across iterations and time. Figure~\ref{fig:res} displays the median and $10-90\%$ percentile of the trajectories.

As expected regarding the previous results on the Hessian spectrum, Picard and Picard-O converge faster than their counterparts relying purely on $\tilde{H}$ as Hessian approximation.

\section*{Conclusion}

This article considers quasi-Newton methods for maximum likelihood ICA using approximated Hessian matrices.
We argue that while the standard Hessian approximation works very well on simulated data, it differs a lot from the true Hessian on most applied problems.  As a consequence, quasi-Newton algorithms which model the curvature of the objective function with such an approximation can have poor convergence rates. We advocate the L-BFGS method to refine  `on the fly' the approximation of the Hessian.  This is supported by experiments on 3 types of real signals which clearly demonstrate that this approach leads to faster convergence.

\clearpage

\bibliographystyle{IEEEtran}
\bibliography{bibliography}

\end{document}